\documentclass[sigconf]{acmart}
\usepackage{subcaption}
\usepackage{dsfont}
\usepackage[linesnumbered,ruled,vlined]{algorithm2e}
\usepackage{xcolor}

\usepackage{multirow}
\usepackage{hyperref}
\usepackage[misc]{ifsym}

\renewcommand\footnotetextcopyrightpermission[1]{}

\acmSubmissionID{7137}

\copyrightyear{2023}
\acmYear{2023}
\setcopyright{acmlicensed}\acmConference[WWW '23]{Proceedings of the ACM Web Conference 2023}{May 1--5, 2023}{Austin, TX, USA}
\acmBooktitle{Proceedings of the ACM Web Conference 2023 (WWW '23), May 1--5, 2023, Austin, TX, USA}
\acmPrice{15.00}
\acmDOI{10.1145/3543507.3583457}
\acmISBN{978-1-4503-9416-1/23/04}

\begin{document}

\title{CEIL: A General Classification-Enhanced Iterative Learning Framework for Text Clustering}



\author{Mingjun Zhao}
\authornote{Both authors contributed equally to this research.}
\affiliation{
  \institution{University of Alberta}
  \city{Edmonton}
  \country{Canada}}
\affiliation{
  \institution{DiDi Global}
  \city{Beijing}
  \country{China}}
\email{zhao2@ualberta.ca}

\author{Mengzhen Wang}
\authornotemark[1]
\affiliation{
  \institution{DiDi Global}
  \city{Beijing}
  \country{China}}
\email{wangmengzhen_i@didiglobal.com	}

\author{Yinglong Ma}
\affiliation{
  \institution{North China Electric Power University}
  \city{Beijing}
  \country{China}}
\email{yinglongma@ncepu.edu.cn}

\author{Di Niu}
\affiliation{
  \institution{University of Alberta}  
  \city{Edmonton}
  \country{Canada}}
\email{dniu@ualberta.ca}

\author{Haijiang Wu$^\text{ \Letter}$}
\affiliation{
  \institution{DiDi Global}
  \city{Beijing}
  \country{China}}
\email{wuhaijiang@didiglobal.com}

\begin{abstract}
Text clustering, as one of the most fundamental challenges in unsupervised learning, aims at grouping semantically similar text segments without relying on human annotations.
With the rapid development of deep learning, deep clustering has achieved significant advantages over traditional clustering methods.
Despite the effectiveness, most existing deep text clustering methods rely heavily on representations pre-trained in general domains, which may not be the most suitable solution for clustering in specific target domains.
To address this issue, we propose \textit{CEIL}, a novel Classification-Enhanced Iterative Learning framework for short text clustering, which aims at generally promoting the clustering performance by introducing a classification objective to iteratively improve feature representations.
In each iteration, we first adopt a language model to retrieve the initial text representations, from which the clustering results are collected using our proposed Category Disentangled Contrastive Clustering (CDCC) algorithm.
After strict data filtering and aggregation processes, samples with clean category labels are retrieved, which serve as supervision information to update the language model with the classification objective via a prompt learning approach.
Finally, the updated language model with improved representation ability is used to enhance clustering in the next iteration.
Extensive experiments demonstrate that the \textit{CEIL} framework significantly improves the clustering performance over iterations, 
and is generally effective on various clustering algorithms.
Moreover, by incorporating \textit{CEIL} on CDCC, we achieve the state-of-the-art clustering performance on a wide range of short text clustering benchmarks outperforming other strong baseline methods.
\end{abstract}




\maketitle

\section{Introduction}
Text clustering, as an unsupervised approach to organizing semantics and analyzing data distributions, plays a vital role in many data-driven natural language processing tasks, such as news recommendation \cite{zheng2013penetrate, lu2014scalable}, topic extraction \cite{zhang2018does}, document summarization \cite{wan2008multi}, etc.

Traditional text clustering approaches transform text into high-dimensional representations based on numerical statistics, such as bag-of-wards (BOW) and term frequency–inverse document frequency (TF-IDF), after which a clustering algorithm is applied to partition text segments into homogeneous groups with respect to a given distance function.
However, in the application of short text clustering, text representations created with traditional measures tend to suffer from sparsity issues, due to the limited length and information within each text piece.

Motivated by the huge success of deep learning in recent years, deep clustering approaches have been increasingly popular.
With the utilization of low-dimensional dense representations extracted from neural networks, deep clustering methods better capture the semantic meaning and alleviate the sparsity issue.
Due to unavailability of supervision in target datasets, unsupervised tasks such as the reconstruction task are often utilized for representation learning with an auto-encoder network \cite{xie2016unsupervised, yang2017towards, hadifar2019self}.

Many studies also leverage the large-scale general domain data for better feature representations.
Word2Vec \cite{mikolov2013distributed} and GloVe \cite{pennington2014glove} learn an embedding vector for each word and are popular choices for text clustering \cite{de2016representation, xu2017self}.
Recently, pre-trained language models (PLMs) such as BERT \cite{devlin2018bert} and GPT \cite{radford2018improving} have achieved great success in improving the representation ability of models, which leads to better clustering results \cite{xu2017self}.
Despite the success, PLMs tend to produce homogeneous representations for all text samples in different clusters, which limits the clustering performance.
Moreover, directly relying on PLM representations may not be optimal due to the gap between the general-purpose corpus on which PLMs are pretrained and the target clustering dataset.

In this paper, we propose \textit{CEIL}, a novel Classification-Enhanced Iterative Learning framework generally applicable to various short text clustering algorithms, in which a classification objective is introduced to improve text representations. The clustering task and the introduced classification task iteratively promote each other, leading to significant performance improvements.
% is introduced to enhance the representations  the clustering task, to iteratively update the feature extraction model for enhanced representations.
% , leading to more precise clustering results.
% \textit{CEIL} is a general framework applicable to various clustering algorithms. 

Specifically, in each iteration, we first extract feature representations with a classification-enhanced language model derived from last iteration.
We use a prompt learning approach to extract text representations, as it can better capture the semantics with limited supervision data compared to the fine-tuning approach.
Then, we propose Category Disentangled Contrastive Clustering (CDCC) algorithm, where a novel category loss is designed to encourage the separation of different clusters, producing better clustering results with more explicit boundaries.
The results are further processed with strict data filtering and aggregation algorithms to alleviate the influence of improperly clustered samples, and are transformed into pseudo classification labels using an automatic verbalizer.
Finally, by learning a classification objective on the clustering dataset using pseudo labels, the language model is updated and acquires better representation ability.

Extensive experiments demonstrate that our \textit{CEIL} framework is generally effective when used with various clustering algorithms, and improves the performance over iterations by exploiting the classification objective to obtain better text representations.
By incorporating \textit{CEIL} on our proposed CDCC algorithm, we  achieve a best average accuracy of $83.2\%$ over 9 short text clustering benchmarks, significantly outperforming other competitive baseline methods by a substantial margin.
% We have also performed comprehensive ablation analysis to validate the effectiveness of each designed component in our proposed framework.

\section{Related Work}
The process of short text clustering mainly involves the derivation of text representations and the clustering framework.
In this section, we first provide a brief overview of text representation methods, and then introduce different clustering frameworks.

\textbf{Text Representations.} Compared with the general text clustering problem, the task of short text clustering is more challenging due to the limited number of words in each text piece, where traditional representation methods such as BoW and TF-IDF often result in sparse vectors with poor expressive ability.
Several studies have addressed this issue by enriching the short text segments with external resources.
\citet{hu2009exploiting,banerjee2007clustering} propose to map the short text segments to Wikipedia articles and leverage the background knowledge for better clustering performance.
In a similar fashion, ontology hierarchical relations and lexical chains are utilized to enhance text representations for clustering \cite{wei2015semantic}.

Low-dimensional representations, such as word embeddings, learned with neural networks have been widely explored and shown great potential to alleviate the sparsity issue.
Different aggregation methods have been explored to produce the sentence representations from word embeddings, including mean-pooling, max-pooling, and weighted summation based on frequency information \cite{de2016representation}.
Although these methods are simple, the word order is lost, which may result in inaccurate semantics.
Recurrent neural networks are exploited to overcome this problem by sequentially modeling the input words and deriving an integrated sentence representation \cite{kiros2015skip}. 

The recent advances of pre-trained language models have achieved great success in building text representations by modeling contextual information \cite{devlin2018bert,radford2018improving}, and shown considerable improvements on short text clustering \cite{zhang2021supporting,li2022simctc}.

\textbf{Clustering Framework.}
% Most of the short text clustering methods adopt a simple framework that first learns the text representations or adopts pre-trained representations, and then apply the clustering algorithm such as K-means \cite{macqueen1967classification,lloyd1982least}.
Most of the studies in short text clustering adopt a simple framework which focuses on the representation learning, and apply clustering algorithms on top of the well-learned representations.
In DEC \cite{xie2016unsupervised} and Self-Train \cite{hadifar2019self}, auto-encoders are employed to solve the reconstruction tasks with the aim of enhancing text representations.
STC$^2$ \cite{xu2017self} incorporates dimensional reduction methods to create auxiliary labels and adopts a convolutional neural network to learn text representations by reconstructing these auxiliary targets.
In SCCL \cite{zhang2021supporting}, authors leverage contrastive learning objectives to promote better separated representations.
Despite the promising performance, there's still room for improvement due to the discrepancy between the representation learning objective and the clustering objective.

Another design idea of clustering framework is to iteratively adjust existing clustering results.
\citet{rakib2020enhancement} propose to train a classification model with cluster labels to correct an initial clustering result by reclassifying the outliers.
However, this method uses fixed representations for clustering, thus may not generalize well.
% This method relies heavily on 
% This method focuses only on adjusting the outliers of an initial clustering result, hence relying heavily on accurate initial representations.

The proposed \textit{CEIL} framework aims to combine the advantages of both designs, by iteratively improving the clustering performance with enhanced text representations, and leveraging the more accurate clusters to promote the representations with the introduced classification objective.
Benefiting from this delicate design, the \textit{CEIL} framework is capable of producing more accurate clusters and steadily improving the clustering performance, compared to previous approaches.

\section{Methodology}

\begin{figure*}[t]
\centering
\includegraphics[scale=0.92]{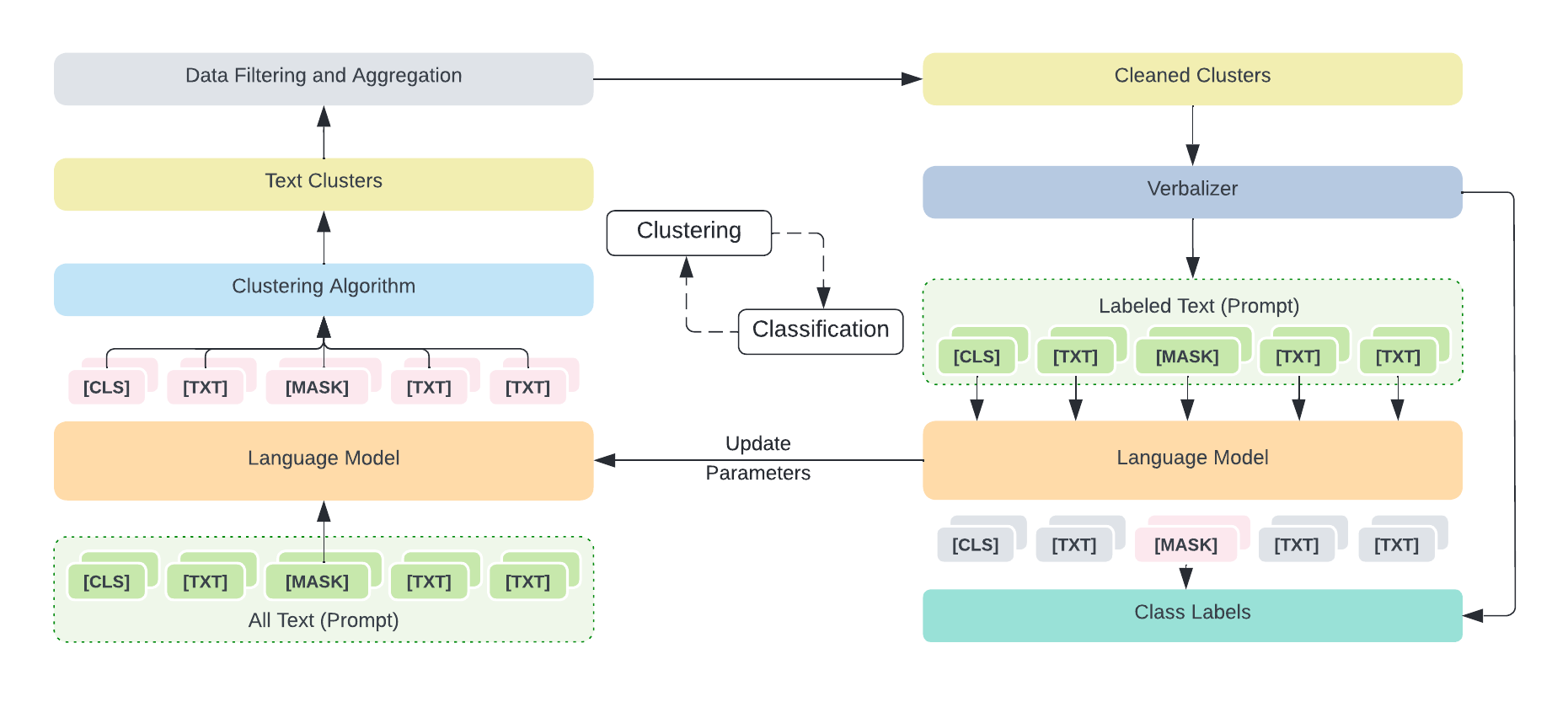}
\caption{
An overview of the proposed Classification-Enhanced Iterative Learning (\textit{CEIL}) framework for short text clustering. 
In each iteration of \textit{CEIL}, the clustering task and the classification task are alternatively optimized.
The clustering exploits the classification-enhanced language model for better text representations and outputs more precise text clusters.
After data filtering and aggregation, the improved clusters are processed with a verbalizer to extract more accurate classification labels, which again enhances the representation ability of the language model by optimizing the classification objective.
}
\label{fig:Overview}
\end{figure*}

In this section, we first provide a general overview of the proposed \textit{CEIL} framework, and introduce our design for the clustering and the classification in detail.

\subsection{Overall Architecture}
Figure \ref{fig:Overview} gives an overview of our proposed framework, in which a clustering objective as well as a classification objective are designed to promote each other iteratively.
In each iteration, the clustering process provides supervisory category labels for classification, and the classification task updates the language model to produce better text representations for clustering.
% The clustering process exploits the meaningful text representations produced by the classification language model to produce accurate text clusters, which are utilized as supervisory signals in the optimization of the classification task.

Specifically, in clustering, we adopt a pre-trained language model at the first iteration, and use the language model trained on the classification task in the following iterations, to extract the text representations.
The prompt-tuning approach is applied during representation extraction, as it is able to produce more semantically meaningful representations especially under the circumstances when the downstream data is limited where little fine-tuning can be performed.
Then, with the extracted representations, we can apply any applicable clustering algorithm, e.g., simple K-means or our proposed CDCC, to derive the text clusters.
The clustering results are cleaned with strict filtering and aggregation to reduce the influence of inaccurate clustering.
Then, the cleaned results are processed with a verbalizer to derive accurate category labels which serve as supervisory signals for the classification task.
% The classified texts are then used to prompt-tune a pre-trained language model via supervised classification, 
By training to classify the text segments into different categories, we effectively enhance the representation ability of the language model, which further benefits the clustering in the next iteration.

\subsection{Category Disentangled Contrastive Clustering}
Contrastive Learning (CL) \cite{wu2018unsupervised,chen2020simple,he2020momentum} aims at learning low-dimensional representations of data where similar samples are pulled together, while dissimilar samples are pushed apart.
% , by pulling together different augmented samples from the same instance, and pushing apart those from different instances.
% As a result, the representations of samples are better separated and is beneficial to clustering.
In SCCL \cite{zhang2021supporting}, an contrastive loss is incorporated along with the clustering loss to encourage better separation of representations.
Despite its effectiveness, the contrastive loss encourages separation of any pair of two samples, disregarding their semantic similarities.
As a potential result, we observe that many samples locate at the halfway of two clusters, leading to ambiguous cluster boundaries.

In order to resolve this issue, we propose Category Disentangled Contrastive Clustering (CDCC) which improves the existing CL-based clustering algorithm by combining the existing contrastive loss and clustering loss with a novel category objective to encourage better separation of different clusters and more explicit cluster boundaries.
% In the following passages, we briefly introduce the three objectives in our clustering algorithm.

\subsubsection{Representation Extraction.}
Either traditional clustering or deep clustering algorithms require the representations of data samples as input.
The quality of initial data representations are crucial to clustering and have great impact on the final performance.

Pre-trained language models show great potential on constructing contextual text representations.
However, without fine-tuning on target data, the expressive ability of the produced representations is limited.
In order to obtain better initial text representations, we adopt the prompt approach for extracting representations by appending a carefully designed template with a [MASK] token, which better summarizes the semantics of the text.
Specifically, in the first iteration of \text{CEIL}, as the language model has not been updated with prompt tuning, we choose the mean vector of all tokens in the prompted input text as the representation, while in the following iterations, the vector of [MASK] produced from the PLM is used as the representation.

\begin{figure*}[t]
\centering
\includegraphics[width=\textwidth]{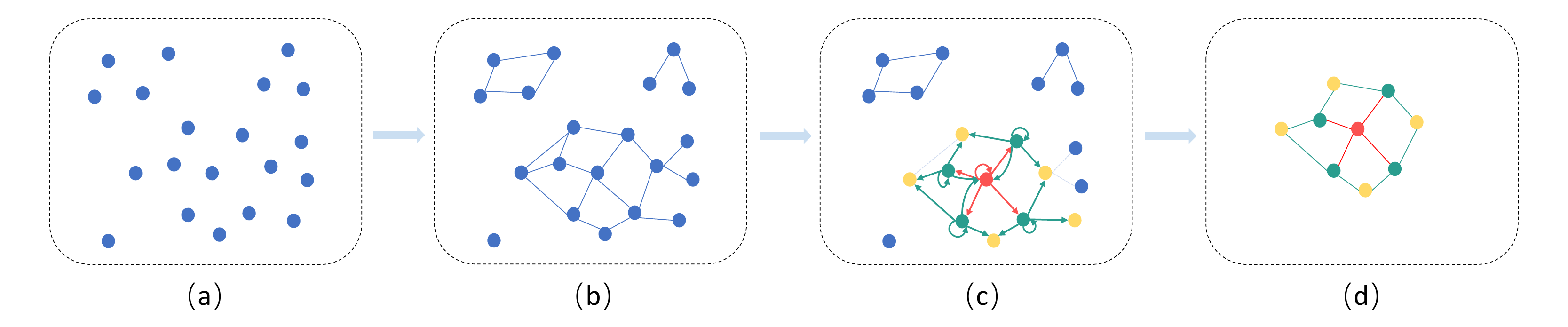}
\caption{
The data filtering process. (a) The sample distribution in a cluster. (b) The generated graph. (c) The search procedure. (d) The filtered cluster.
}
\label{fig:data_filtering}
\end{figure*}

\subsubsection{Contrastive Loss.} 
Specifically, given a mini-batch of $M$ input text segments $\mathcal{B}=\{x_k\}_{k=1}^M$, 
an augmented batch $\mathcal{B}^a=\{\tilde{x}_k\}_{k=1}^{2M}$ is constructed by applying two random augmentations on each sample $x_k$ to derived augmented samples $\tilde{x}_{2k-1}$ and $\tilde{x}_{2k}$ as a positive pair in contrastive learning.
The other samples in $\mathcal{B}^a$ are treated as negative instances regarding this positive pair.
% Any sample in the augmented batch $\mathcal{B}^a$ other than $\tilde{x}_{2i+1}$ is treated as negetive samples to $\tilde{x}_{2i}$.
Let $\tilde{h}_i=\phi(\tilde{x}_i)$ be the representation of sample $\tilde{x}_i$ extracted with language model $\phi$.
The loss function for a positive pair of samples $(i,j)$ is defined as 
\begin{equation}
    \ell(i,j) = -\text{log} \dfrac{\text{exp}(\text{cos}(\tilde{h}_{i}, \tilde{h}_{j})/\tau)}{\sum_{k=1}^{2M} \mathds{1}_{k\neq i} \cdot \text{exp}(\text{cos}(\tilde{h}_{i}, \tilde{h}_{k})/\tau)},
\end{equation}
where $\text{cos}(\cdot)$ refers to the cosine similarity function.
The contrastive loss is derived by calculating the average loss on all samples in the augmented batch $\mathcal{B}^a$:
\begin{equation}
    \mathcal{L}_\text{contrast} = \dfrac{1}{2M} \sum_{i=1}^{M} \big[\ell(2k-1,2k) + \ell(2k,2k-1)\big].
\end{equation}

\subsubsection{Clustering Loss.}
The clustering loss is first proposed in DEC \cite{xie2016unsupervised}, and adopted in short text clustering \cite{hadifar2019self, zhang2021supporting}.
It aims to improve the clustering by iterating between computing an auxiliary target distribution and minimizing the Kullback-Leibler (KL) divergence to it.

Given a text sample $x_i$ in mini-batch $\mathcal{B}$, its representation $h_i$, and the learnable centroid vector $\mu_j$ of cluster $j$, we calculate the probability $q_{ij}$ of assigning $x_i$ to cluster $j$ with the Student's \textit{t}-distribution:
\begin{equation}
\label{eq:prob}
    q_{ij} = \dfrac{(1+\rVert h_i-\mu_j\rVert_2^2/\alpha)^{-\frac{\alpha+1}{2}}}{\sum_{j'=1}^K (1+\rVert h_i-\mu_{j'}\rVert_2^2/\alpha)^{-\frac{\alpha+1}{2}}},
\end{equation}
where $K$ is the number of clusters and $\alpha$ denotes the degree of freedom.
Following \citet{van2008visualizing}, we set $\alpha$ to 1.

The auxiliary distribution $p_{ij}$ is computed as:
\begin{equation}
    p_{ij} = \dfrac{q_{ij}^2/f_i}{\sum_{j'}^Kq_{ij}^2/f_{j'}},
\end{equation}
where $f_j = \sum_i^M q_{ij}$ is the soft cluster frequencies and is approximated within the $M$ samples in mini-batch $\mathcal{B}$.

The clustering loss is defined to jointly optimize the cluster centroids $\{\mu_j\}_{j=1}^K$ and the language model $\phi$ by minimizing the KL-divergence between the soft assignments $q_i$ and the auxiliary distribution $p_i$ as:
\begin{equation}
\begin{aligned}
    \mathcal{L}_\text{cluster} = \sum_{i=1}^M \text{KL}\big[p_i\rVert q_i\big] = \sum_{i=1}^M \sum_{j=1}^K p_{ij} \text{log} \frac{p_{ij}}{q_{ij}}.
\end{aligned}
\end{equation}

\subsubsection{Category Loss.}
One critical problem of clustering is that the initial data representations have significant overlap across categories.
Although the fore-mentioned contrastive learning objective alleviates this issue by separating individual samples in the representation space, we still observe the presence of many sample points located at the shared boundary of two clusters, potentially leading to incorrect clustering and ambiguous cluster boundaries.
To resolve this issue, we propose a novel category disentangling objective which aims at improving clustering by introducing more room between clusters and obtaining more explicit cluster boundaries.

Based on the probability $q_{ij}$ calculated with Equation \ref{eq:prob}, we first make hard assignments for each sample $x_i$ to cluster $c_k$, where $k=\text{argmax}_j(q_{ij})$.
Then, for each cluster $c_k$, we obtain its initial representation vector $h_{c_k}=\text{avg}(\{x_i|x_i \in c_k\})$ by taking the average of the representations of samples in it.

The initial representation can be inaccurate especially during the first few iterations when the clustering results are not very stable.
Therefore, we apply a simple threshold $\theta$ to obtain a filtered cluster $c_k'=\{x_i|\text{cos}(h_i,h_{c_k})\ge\theta,x_i\in c_k\}$.
Similarly, for each filtered cluster $c_k'$, we derive a final representation vector $h_{c_k'}=\text{avg}(\{x_i|x_i \in c_k'\})$.
With the aim of better separating the clusters, the category loss is proposed to push apart the cluster feature representations by maximizing the cosine distance between clusters, which is defined as:
\begin{equation}
    \mathcal{L}_\text{category}=\text{exp}\big(\dfrac{\sum_{i=1}^{K-1} \sum_{j=i+1}^K \text{cos}(h_{c_i'},h_{c_j'})}{(K-1)(K-2)/2}\big).
\end{equation}

\subsubsection{Overall Objective.}
In summary, the overall objective of CDCC is:
\begin{equation}
    \mathcal{L}=\mathcal{L}_\text{contrast}+ \lambda \mathcal{L}_\text{cluster} + \mathcal{L}_\text{category},
\end{equation}
where $\lambda$ is a loss coefficient to balance the objectives.

\subsection{Data Filtering and Aggregation}
After clustering, each sample $x_i$ is assigned a category label $c_k$, which can later be utilized as the supervisory label for the classification objective.
However, the results produced by clustering algorithm can hardly be perfect, where samples from multiple ground truth categories can be allocated into one cluster, and samples from the same category may be split into multiple clusters.
Directly incorporating the unprocessed clustering results as training data for classification may introduce much noise and jeopardize the performance of the classifier.
Therefore, we propose to incorporate a data filtering as well as a data aggregation module to alleviate this problem.

\subsubsection{Data Filtering.}
We propose a data filtering algorithm for cleaning clustering results by mining the adjacency relationships between sample points in the same cluster and keeping a collection of samples with abounding inter-relations as the filtered cluster.
% The outline of this filtering algorithm is described as follows. First, we find the sample $x_i$ in the cluster $c_k$ that has the most neighbours, where we define a sample $x_j$ is a neighbor of $x_i$ if their similarity is greater than a threshold $\beta$:
% \begin{equation}
%     \text{neighbor}(x_i) = \{x_j | \text{cos}(x_i, x_j) \ge \beta, x_j \in c_k\}.
% \end{equation}
% Then, we keep a collection containing $x_i$, its neighbours $\text{neighbor}(x_i)$, and its neighbours' neighbours $\{neighbor(x_j) | x_j \in neighbor(x_i)\}$.
% We also keep track of the number of neighbors for each sample.
% Finally, for the filtered cluster, we gather the statistics and select samples from the collection whose number of neighbors is greater than or equal to the mode number of neighbors.

The data filtering process is illustrated in Figure \ref{fig:data_filtering}, where the sample distribution of a cluster is shown in (a). 
In (b), we create a graph where the nodes represent the samples and the edges between nodes represent the adjacency relationship, which are created when the cosine similarity of two nodes is greater than a threshold $\beta$.
The search procedure is demonstrated in (c).
Starting from the center node colored in red with the most connections, we retrieve its adjacent neighbors colored in green and itself through the red arrows.
The same process is repeated on all green nodes to retrieve the second-order neighbors colored in yellow through the green arrows. 
We take all the retrieved nodes as candidates, gather the statistics of their input degrees, i.e., the number of arrows pointing to a node, and select the samples whose input degree is greater than the mode value as the filtered cluster as shown in (d).

This data filtering process guarantees only tightly clustered samples remain, which efficiently alleviate the noise that could be potentially introduced to the classification task.

\subsubsection{Data Aggregation.}
This process aims to merge clusters close enough to each other which are very likely to come from the same ground truth category.
We adopt a simple aggregation strategy by measuring the similarity between two clusters $c_i'$ and $c_j'$, and merging these two clusters if their similarity is greater than a threshold $\delta$.
The merge will eliminate the original two clusters and produce a new cluster $c_{ij}'' = c_i' \cup c_j'$.

Empirically, the data aggregation proves to be effective in improving the clustering performance especially on imbalanced datasets with plenty of categories.

\subsection{Classifier Verbalizer}

In prompt learning, the verbalizer is an important module that bridges a projection between the vocabulary and class labels, and has critical impacts on the classification performance.
Most existing works use manual verbalizers, where specific tokens are selected by human for each class, which is not only time-consuming and requiring expert knowledge, but also not applicable to our clustering circumstances, where no supervision is available.

\begin{figure}[t]
\centering
\includegraphics[width=\columnwidth]{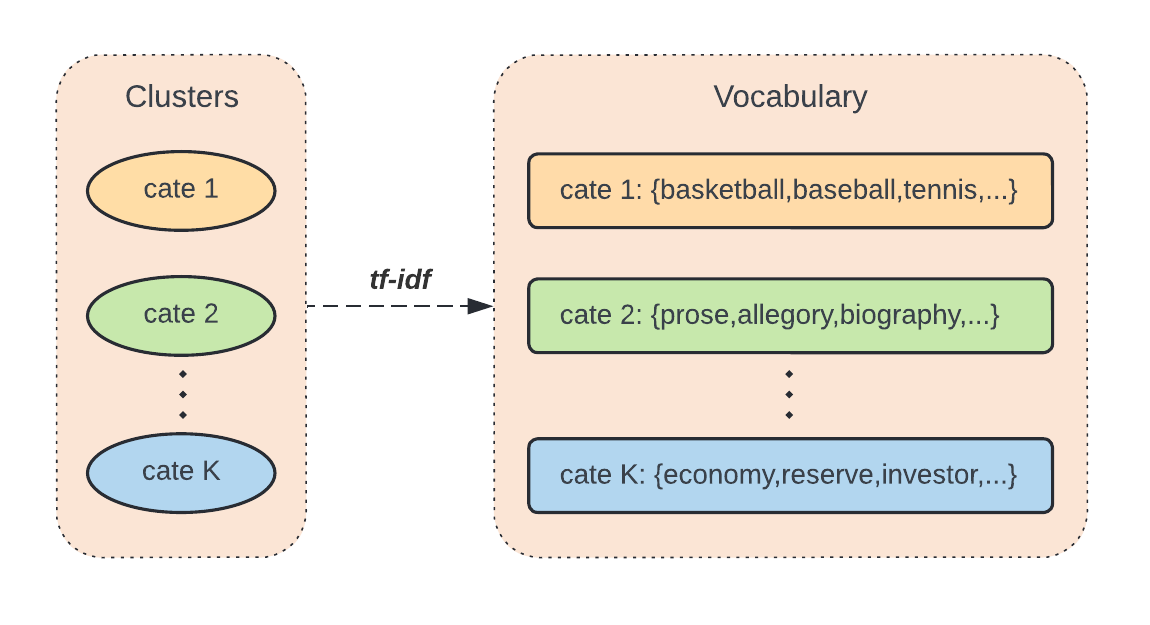}
\caption{
Our automatic verbalizer maps the cluster categories to correlated tokens in the vocabulary based on token frequencies.
}
\label{fig:Verbalizer}
\end{figure}

As shown in Figure \ref{fig:Verbalizer}, we propose an automatic verbalizer to map meaningless category labels in clustering to concrete representative tokens in the vocabulary.
Specifically, for each cleaned cluster $c_i''$, we gather the token frequencies and extract a collection of $n$ key words $\mathcal{V}_{c_i''}=\{t_k\}^n$ based on the TF-IDF value of each token in the vocabulary $\mathcal{V}$.
The extracted keywords $\mathcal{V}_{c_i''}$ are then utilized as the supervisory labels for the classification task.

\subsection{Classification}
Inspired by \cite{yang2016joint}, we use the classification objective to update the language model $\phi$ for better representation learning, and further improve the clustering performance.

Consistent with the update of language model in clustering, we also adopt prompt learning for the optimization of the classification task.
Given a prompted text sample $x_j$ in cluster $c_i''$ and the corresponding classification labels $\mathcal{V}_{c_i''}$, the objective is to maximize the probability of the label $y_j$ calculated with the language model $\phi$ by averaging the probabilities of tokens in $\mathcal{V}_{c_i''}$ filled in the [MASK] token: 
\begin{equation}
\begin{aligned}
    &\max_\phi P_\phi(y_j|x_j) \\
    = &\max_\phi \Big[ \frac{1}{|\mathcal{V}_{c_i''}|} \sum_{t\in \mathcal{V}_{c_i''}} \big( P_\phi(\text{[MASK]}=t|x_j)  \big) \Big],
\end{aligned}
\end{equation}
where the update of the language model $\phi$ is achieved by gradient descent.

By utilizing the strictly cleaned clustering results as the supervisory signals, the representation ability of the language model is further improved by learning on the classification objective to capture the semantics of the text.  
The updated language model is then exploited in the next iteration for better text representations, leading to more accurate clusters and classification labels of higher quality, which again improves the language model.

% The iterative update of our \textit{CEIL} framework 
With the iterative update on the clustering and classification objectives, and the fore-mentioned delicately designed modules, our \textit{CEIL} framework learns better text representations and significantly improves clustering performance.

\section{Experiments}
In this section, we will first describe the datasets used in our evaluation and provide the implementation details along with the performance results.

\begin{table}[tb]
    \centering
    \caption{Dataset statistics. $|\mathcal{V}|$: the vocabulary size; N: number of short text documents; L: average number of words in each document; C: number of clusters; P: the ratio of the size of the largest cluster to that of the smallest cluster.}
    \begin{tabular}{cccccc}
        \toprule
        Dataset & $|\mathcal{V}|$ & N & L & C & P \\
        \midrule
        AgNews & 21K & 8000 & 23 & 4 & 1 \\
        StackOverflow & 15K & 20000 & 8 & 20 & 1 \\
        Biomedical & 19K & 20000 & 13 & 20 & 1 \\
        SearchSnippets & 31K & 12340 & 18 & 8 & 7 \\
        GoogleNews-TS & 20K & 11109 & 28 & 152 & 143 \\
        GoogleNews-S & 18K & 11109 & 22 & 152 & 143 \\
        GoogleNews-T & 8K & 11109 & 6 & 152 & 143 \\
        Tweet & 5K & 2472 & 8 & 89 & 249 \\
        MX & 17K & 20000 & 26 & 210 & 15 \\
        \bottomrule
    \end{tabular}
    \label{tab:dataset}
\end{table}

\begin{table*}[tb]
    \centering
    % \resizebox{\columnwidth}{!}{
    \caption{The hyperparameters and prompt templates adopted for different tasks in our experiment.}
    \begin{tabular}{ccccccccl}
        \toprule
        Dataset & $lr$ & bs$_\text{cluster}$ & bs$_\text{classifier}$ & $\theta$ & $\beta$ & $\delta$ & $n$ & Prompt Template\\
        \midrule
        AgNews & 1e-5 & 100 & 64 & 0.7 & 0.93 & 1 & 15 & [X] This topic is about [MASK] .\\
        StackOverflow & 1e-5 & 400 & 300 & 0.7 & 0.6 & 1 & 24 & [X] [MASK] .\\
        Biomedical & 1e-5 & 100 & 64 & 0.7 & 0.6 & 1 & 20 & [MASK] [X] .\\
        SearchSnippets & 1e-5 & 100 & 64 & 0.7 & 0.6 & 1 & 25 & [X] is [MASK] .\\
        GoogleNews-TS & 5e-6 & 100 & 64 & 0.7 & 0.8 & 0.98 & 9 & [MASK] [X] .\\
        GoogleNews-S & 1e-5 & 100 & 64 & 0.7 & 0.76 & 0.99 & 39 & [X] This topic is about [MASK] .\\
        GoogleNews-T & 5e-6 & 400 & 300 & 0.7 & 0.78 & 0.98 & 35 & Category : [MASK] [X]\\
        Tweet & 1e-5 & 100 & 64 & 0.7 & 0.8 & 0.95 & 10 & [MASK] [X] . \\
        MX & 1e-5 & 100 & 64 & 0.7 & 0.6 & 0.95 & 10 & This food is made up of [MASK] .\\
        \bottomrule
    \end{tabular}
    % }
    \label{tab:hyperparameter}
\end{table*}

\subsubsection{Datasets.}
Our experiments are based on eight public short text clustering datasets and an internal dataset MX. Table \ref{tab:dataset} summarizes the statistics of the evaluation datasets.
% Eight public short text clustering datasets are used for evaluation, i.e., AgNews \cite{zhang2015text,rakib2020enhancement}, StackOverflow \footnote{https://www.kaggle.com/c/predict-closed-questions-on-stack-overflow/}, Biomedical \footnote{http://participants-area.bioasq.org}, SearchSnippets \cite{phan2008learning}, GoogleNews-TS, GoogleNews-S, GoogleNews-T \cite{yin2016model,rakib2020enhancement}, Tweet \cite{yin2016model}, and MX. 

% Due to the limited space, we introduce them in Appendix.

\begin{itemize}
    \item \textbf{AgNews} \cite{zhang2015text} is a dataset of news articles. The news titles of 4 topics selected by HAC-SD \cite{rakib2020enhancement} are used as the short text segments for clustering. 
    \item \textbf{StackOverflow} \footnote{https://www.kaggle.com/c/predict-closed-questions-on-stack-overflow/} contains 20,000 question titles associated with 20 categories from StackOverflow websites, selected by STCC \cite{xu2017self}. 
    \item \textbf{Biomedical} \footnote{http://participants-area.bioasq.org} is a subset of the PubMed data distributed by BioASQ. It contains randomly selected 20,000 page titles from 20 groups by STCC \cite{xu2017self}.
    \item \textbf{SearchSnippets} \cite{phan2008learning} is a dataset extracted from web search snippets, consists of 12,340 snippets with 8 groups.
    \item \textbf{GoogleNews} \cite{yin2016model} contains news titles and snippets from 11,109 articles associated with 152 events. Following HAC-SD \cite{rakib2020enhancement}, GoogleNews-TS, GoogleNews-T, and GoogleNews-S are three subsets containing titles and snippets, titles only, and snippets only, respectively.
    \item \textbf{MX} is an internal dataset on dish recipes in Spanish, consists of more than 3M text in 150 course-grained categories, which are split into fine-grained clusters. Table \ref{tab:dataset} gives the average statistics on different categories.
\end{itemize}

% \begin{table}[tb]
%     \centering
%     \begin{tabular}{c|l}
%         \toprule
%         Dataset & Prompt Template \\
%         \midrule
%         AgNews & [X] This topic is about [MASK] .\\
%         StackOverflow & [X] [MASK] .\\
%         Biomedical & [MASK] [X] .\\
%         SearchSnippets & [X] is [MASK] .\\
%         GoogleNews-TS & [MASK] [X] .\\
%         GoogleNews-S & [X] This topic is about [MASK] .\\
%         GoogleNews-T & Category : [MASK] [X]\\
%         Tweet & [MASK] [X] . \\
%         MX & This food is made up of [MASK] .\\
%         \bottomrule
%     \end{tabular}
%     \caption{Prompt template design used in different datasets.}
%     \label{tab:prompt_template}
% \end{table}

\begin{table*}[tb]
    \caption{Comparison results of different short text clustering methods on nine evaluation datasets. The results of applying our proposed \textit{CEIL} framework on GMM, SCCL, and CDCC as GMM\_CEIL, SCCL\_CEIL, and CDCC\_CEIL are also included.}
    \begin{subtable}[h]{\textwidth}
    \centering
    \caption{Accuracy and NMI results.}
    \resizebox{\textwidth}{!}{
    \begin{tabular}{c|cc|cccccccccccccccccc}
        \toprule
        \multirow{2}{*}{Method} & \multicolumn{2}{c|}{Average} & \multicolumn{2}{c}{AgNews} & \multicolumn{2}{c}{StackOverflow} & \multicolumn{2}{c}{Biomedical} & \multicolumn{2}{c}{SearchSnippets} & \multicolumn{2}{c}{Tweet} & \multicolumn{2}{c}{GoogleNews-TS} & \multicolumn{2}{c}{GoogleNews-S} & \multicolumn{2}{c}{GoogleNews-T} & \multicolumn{2}{c}{MX} \\
         & ACC & NMI & ACC & NMI & ACC & NMI & ACC & NMI & ACC & NMI & ACC & NMI & ACC & NMI & ACC & NMI & ACC & NMI & ACC & NMI \\
        \midrule
        BOW & - & - & $27.6$ & $2.6$ & $18.5$ & $14.0$ & $14.3$ & $9.2$ & $24.3$ & $9.3$ & $49.7$ & $73.6$ & $57.5$ & $81.9$ & $49.0$ & $73.5$ & $49.8$ & $73.2$ & - & - \\
        TF-IDF & - & - & $34.5$ & $11.9$ & $58.4$ & $58.7$ & $28.3$ & $23.2$ & $31.5$ & $19.2$ & $57.0$ & $80.7$ & $68.0$ & $88.9$ & $61.9$ & $83.0$ & $58.9$ & $79.3$ & - & - \\
        STCC & - & - & - & - & $51.1$ & $49.0$ & $43.6$ & $38.1$ & $77.0$ & $63.2$ & - & - & - & - & - & - & - & - & - & - \\
        Self-Train & - & - & - & - & $59.8$ & $54.8$ & $\mathbf{54.8}$ & $\mathbf{47.1}$ & $77.1$ & $56.7$ & - & - & - & - & - & - & - & - & - & - \\
        HAC\_SD & $76.4$ & $69.3$ & $81.8$ & $54.6$ & $64.8$ & $59.5$ & $40.1$ & $33.5$ & $82.7$ & $63.8$ & $89.6$ & $85.2$ & $85.8$ & $88.0$ & $80.6$ & $83.5$ & $81.8$ & $84.2$ & $80.7$ & $71.6$ \\
        HAC\_SD\_IC & $82.2$ & $75.2$ & $84.5$ & $59.1$ & $78.7$ & $73.4$ & $47.8$ & $41.3$ & $87.7$ & $71.9$ & $\mathbf{91.5}$ & $86.9$ & $92.3$ & $93.2$ & $\mathbf{89.0}$ & $90.0$ & $\mathbf{87.2}$ & $87.9$ & $81.5$ & $72.8$ \\
        GMM & $58.7$ & $60.2$ & $68.0$ & $33.4$ & $57.3$ & $49.2$ & $38.9$ & $32.9$ & $56.3$ & $32.5$ & $52.8$ & $78.1$ & $64.4$ & $86.7$ & $57.5$ & $80.7$ & $54.9$ & $78.1$ & $78.1$ & $69.8$ \\
        GMM\_CEIL & $76.2$ & $74.6$ & $86.6$ & $64.7$ & $75.9$ & $74.7$ & $44.0$ & $43.9$ & $77.9$ & $58.9$ & $87.8$ & $92.3$ & $84.3$ & $92.8$ & $77.9$ & $88.1$ & $70.9$ & $84.8$ & $80.6$ & $71.2$ \\
        SCCL & $78.1$ & $76.8$ & $88.2$ & $68.2$ & $75.5$ & $74.5$ & $46.2$ & $41.5$ & $85.2$ & $71.1$ & $78.2$ & $89.2$ & $89.8$ & $94.9$ & $83.1$ & $90.4$ & $75.8$ & $88.3$ & $80.6$ & $73.2$ \\
        SCCL\_CEIL & $81.0$ & $77.9$ & $88.9$ & $69.2$ & $77.8$ & $76.1$ & $47.3$ & $39.9$ & $88.3$ & $73.9$ & $87.3$ & $92.6$ & $91.3$ & $95.1$ & $85.5$ & $91.6$ & $79.6$ & $89.1$ & $83.2$ & $74.0$ \\
        CDCC & $78.8$ & $76.8$ & $88.3$ & $68.2$ & $76.9$ & $75.3$ & $42.8$ & $38.1$ & $85.3$ & $71.2$ & $82.6$ & $90.2$ & $90.2$ & $94.9$ & $84.5$ & $90.6$ & $76.2$ & $88.5$ & $82.8$ & $74.3$ \\
        CDCC\_CEIL & $\mathbf{83.2}$ & $\mathbf{79.2}$ & $\mathbf{89.1}$ & $\mathbf{69.7}$ & $\mathbf{78.8}$ & $\mathbf{77.3}$ & $49.2$ & $41.0$ & $\mathbf{89.3}$ & $\mathbf{75.2}$ & $91.2$ & $\mathbf{94.1}$ & $\mathbf{92.5}$ & $\mathbf{95.8}$ & $87.2$ & $\mathbf{92.4}$ & $84.3$ & $\mathbf{91.3}$ & $\mathbf{87.3}$ & $\mathbf{76.1}$ \\
        \bottomrule
    \end{tabular}
    }
    \label{tab:acc_nmi}
    \end{subtable}
    
    \begin{subtable}[h]{\textwidth}
    \caption{Standard deviations results.}
    \centering
    \resizebox{\textwidth}{!}{
    \begin{tabular}{c|cccccccccccccccccc}
        \toprule
        \multirow{3}[2]{*}{Method} & \multicolumn{2}{c}{AgNews} & \multicolumn{2}{c}{StackOverflow} & \multicolumn{2}{c}{Biomedical} & \multicolumn{2}{c}{SearchSnippets} & \multicolumn{2}{c}{Tweet} & \multicolumn{2}{c}{GoogleNews-TS} & \multicolumn{2}{c}{GoogleNews-S} & \multicolumn{2}{c}{GoogleNews-T} & \multicolumn{2}{c}{MX} \\
        % \cline{2-19}
        \cmidrule(lr){2-19}
        & \vtop{\hbox{\strut ACC}\hbox{\strut \_std}} & \vtop{\hbox{\strut NMI}\hbox{\strut \_std}} & \vtop{\hbox{\strut ACC}\hbox{\strut \_std}} & \vtop{\hbox{\strut NMI}\hbox{\strut \_std}} & \vtop{\hbox{\strut ACC}\hbox{\strut \_std}} & \vtop{\hbox{\strut NMI}\hbox{\strut \_std}} & \vtop{\hbox{\strut ACC}\hbox{\strut \_std}} & \vtop{\hbox{\strut NMI}\hbox{\strut \_std}} & \vtop{\hbox{\strut ACC}\hbox{\strut \_std}} & \vtop{\hbox{\strut NMI}\hbox{\strut \_std}} & \vtop{\hbox{\strut ACC}\hbox{\strut \_std}} & \vtop{\hbox{\strut NMI}\hbox{\strut \_std}} & \vtop{\hbox{\strut ACC}\hbox{\strut \_std}} & \vtop{\hbox{\strut NMI}\hbox{\strut \_std}} & \vtop{\hbox{\strut ACC}\hbox{\strut \_std}} & \vtop{\hbox{\strut NMI}\hbox{\strut \_std}} & \vtop{\hbox{\strut ACC}\hbox{\strut \_std}} & \vtop{\hbox{\strut NMI}\hbox{\strut \_std}} \\
        \midrule
        GMM & 0.66 & 0.58 & 1.30 & 0.61 & 0.65 & 1.31 & 1.6 & 0.72 & 1.30 & 0.66 & 0.11 & 0.22 & 0.10 & 0.12 & 0.69 & 0.50 & 0.40 & 0.13 \\
        GMM\_CEIL & 0.52 & 0.58 & 0.92 & 0.51 & 0.21 & 0.31 & 0.66 & 0.78 & 0.98 & 0.78 & 0.17 & 0.09 & 0.61 & 0.55 & 0.71 & 0.23 & 0.36 & 0.28 \\
        CDCC & 0.23 & 0.54 & 1.70 & 2.05 & 0.17 & 2.4 & 1.30 & 1.40 & 0.61 & 0.39 & 0.48 & 0.26 & 0.33 & 0.32 & 0.70 & 0.32 & 0.21 & 0.10 \\
        CDCC\_CEIL & 0.17 & 0.33 & 0.70 & 0.33 & 1.20 & 0.57 & 0.60 & 1.03 & 1.96 & 0.78 & 0.34 & 0.08 & 1.27 & 0.18 & 0.83 & 0.26 & 0.21 & 0.13 \\
        \bottomrule
    \end{tabular}
    }
    \label{tab:std}
    \end{subtable}
    \label{tab:Compare}
\end{table*}

\subsubsection{Evaluation Metrics.}
Following the common practice, we evaluate the performance of different short text clustering methods via top-1 accuracy (ACC) and Normalized Mutual Information (NMI).

\subsubsection{Baselines.}
We compare our proposed method with traditional text clustering methods and a number of state-of-the-art methods, including BoW, TF-IDF, GMM, Self-Train \cite{hadifar2019self}, HAC-SD \cite{rakib2020enhancement}, and STCC \cite{zhang2021supporting}.
% An introduction of these methods is given in Appendix.
\begin{itemize}
    \item \textbf{BoW \& TF-IDF} features are extracted with a feature dimension of 1,500, and are incorporated with K-means clustering.
    \item \textbf{GMM} is a traditional clustering method using Gaussian Mixture Models.
    \item \textbf{STCC} \cite{xu2017self} incorporates Word2Vec \cite{mikolov2013distributed} word embeddings and a convolutional neural network to construct the text representation, and applies K-means for clustering.
    \item \textbf{Self-Train} \cite{hadifar2019self} enhances word embeddings using SIF \cite{arora2017simple}, and adopts an auto-encoder for representations. 
    \item \textbf{HAC\_SD} \cite{rakib2020enhancement} uses hierarchical agglomerative clustering clustering on a sparse pairwise similarity matrix. \textbf{HAC\_SD\_IC} further updates the clustering results by iteratively reclassifying the outliers.
\end{itemize}

\subsubsection{Implementation Details.}
We implement our \textit{CEIL} framework based on PyTorch \cite{paszke2017automatic} and OpenPrompt \cite{ding2021openprompt}, and incorporate \textit{distilbert-base-nli-stsb-mean-tokens} from Sentence Transformer \cite{reimers2019sentence} as the initial checkpoint of the language model.
In the application of \textit{CEIL} framework, we adopt $5$ iterations of update when applied on SCCL and CDCC, and $8$ iterations on GMM.
% For different datasets, we use grid search to choose hyperparameters, and list the adopted hyperparameters and prompt templates in Appendix.

In Table \ref{tab:hyperparameter}, we list the adopted hyperparameters and prompt templates for each dataset.
We choose the learning rate between 1e-5 and 5e-6, the batch size of clustering between 100 and 400, and the batch size of classifier between 64 and 300.
The loss coefficient for the clustering loss $\lambda$ is set to 1.
The threshold $\theta$ used in the category loss is set to $0.7$.
For the choice of filtering threshold $\beta$, and aggregation threshold $\delta$, we use a grid search with an interval of $0.01$ from $0.6$ to $1.0$ and $0.95$ to $1.0$ respectively.
And the number of keyword $n$ in verbalizer is also chosen by grid search with an interval of $1$.
For each dataset, we adopt the best prompt template selected from the classic manual designed templates from OpenPrompt \cite{ding2021openprompt} by a simple K-means experiment.

\subsection{Comparison with State-of-the-art Baselines}
We conduct comprehensive experiments on the evaluation datasets to demonstrate the effectiveness of our proposed CDCC clustering algorithm and the \textit{CEIL} framework.
% make a comprehensive comparison of our proposed CDCC clustering algorithm and the \textit{CEIL} framework, with other baseline methods.
We have also proved the general applicability of the \textit{CEIL} framework by applying it on existing text clustering algorithms including GMM, SCCL and CDCC.
For evaluation, we conduct five repetitive runs and report the mean performance in Table \ref{tab:acc_nmi}, and include the standard deviation results in \ref{tab:std}.
% We run five repetitive experiments and report the mean performance.
% The evaluation results are summarized in Table \ref{tab:Compare}.
% Table \ref{tab:Compare} summarizes the results on the evaluation datasets.

From the results we can see that among the clustering algorithms without iterative updates, our CDCC achieves the best average performance of $78.8\%$ ACC and $76.8\%$ NMI, outperforming the competitive baselines including SCCL \cite{zhang2021supporting} and HAC\_SD \cite{rakib2020enhancement}.
Note that HAC\_SD\_IC also adopts iterative updates by iteratively reclassifying the outliers.
Out of 9 evaluation tasks, our CDCC beats SCCL in 8 tasks, demonstrating that our proposed category loss can effectively obtain better separation of clusters and generally improve the clustering performance under different circumstances.

The \textit{CEIL} framework also shows promising effects on boosting clustering performance.
By comparing the \text{CEIL} enhanced methods GMM\_CEIL, SCCL\_CEIL and CDCC\_CEIL with their original version, we can see that applying \textit{CEIL} framework stably improves the clustering performance on all evaluation tasks, and brings significant performance gains of $17.5\%$, $2.9\%$ and $4.4\%$ average accuracy improvements on the traditional GMM method, and more advanced deep clustering methods, SCCL and CDCC, respectively.  
Moreover, among all clustering methods, our CDCC\_CEIL achieves the top accuracy on 5 out of 9 tasks and the best NMI on 8 out of 9 tasks, while obtaining the best average performance of $83.2\%$ ACC and $79.2\%$ NMI.
% \blue{Note that on Biomedical dataset, CDCC\_CEIL only achieves the second best performance because Self-Train \cite{hadifar2019self} incorporates additional pre-training on a in-domain large-scale }
This result provides strong evidence that the proposed \textit{CEIL} framework can produce better text clusters by iteratively promoting the text representations with the classification objective, and is generally effective to various text clustering algorithms. 

We have also conducted thorough analysis to examine the reason why Self-Train \cite{hadifar2019self} has a clear advantage over other methods on the Biomedical dataset.
% excellent performance of Self-Train on the Biomedical dataset, 
% CDCC\_CEIL does not ace on the \red{Biomedical} and the potential improvements to our method.
The Biomedical dataset contains professional domain-specific knowledge which are hardly covered in general pre-training corpus, while Self-Train \cite{hadifar2019self} incorporates an additional large-scale in-domain corpus to pre-train the word embeddings, leading to better clustering performance.
% Therefore, the additional pre-training on a large-scale in-domain corpus Self-Train \cite{hadifar2019self} than our representations extracted from a general pre-trained language model.
From this observation, we can infer a potential improvement to our method of conducting continual pre-training on in-domain datasets to capture domain-specific knowledge before clustering.

\begin{figure}[t]
\centering
\begin{subfigure}[b]{.49\columnwidth}
  \includegraphics[width=\columnwidth]{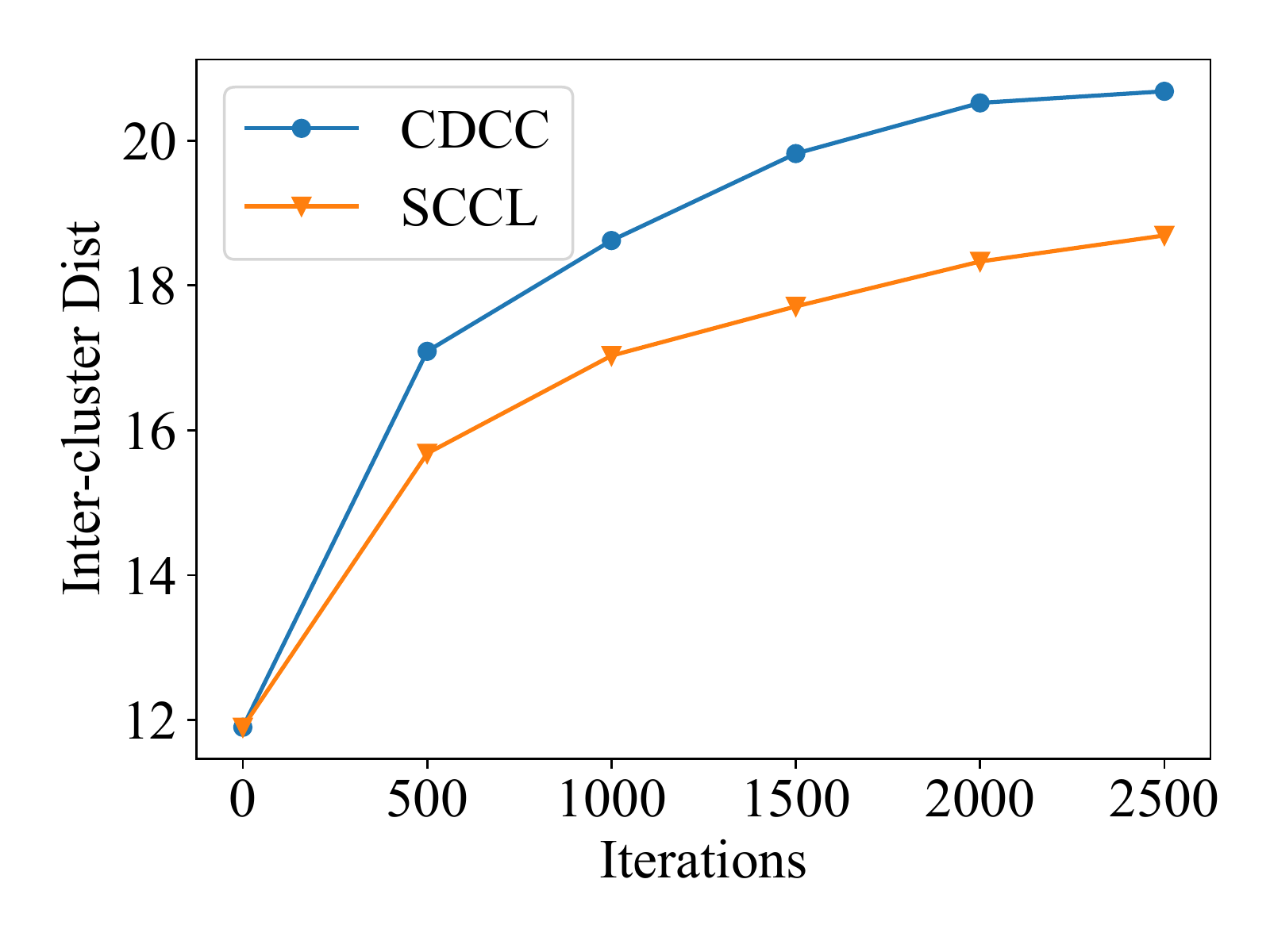}
  \caption{Inter-cluster Distances}
  \end{subfigure}
\begin{subfigure}[b]{.49\columnwidth}
  \includegraphics[width=\columnwidth]{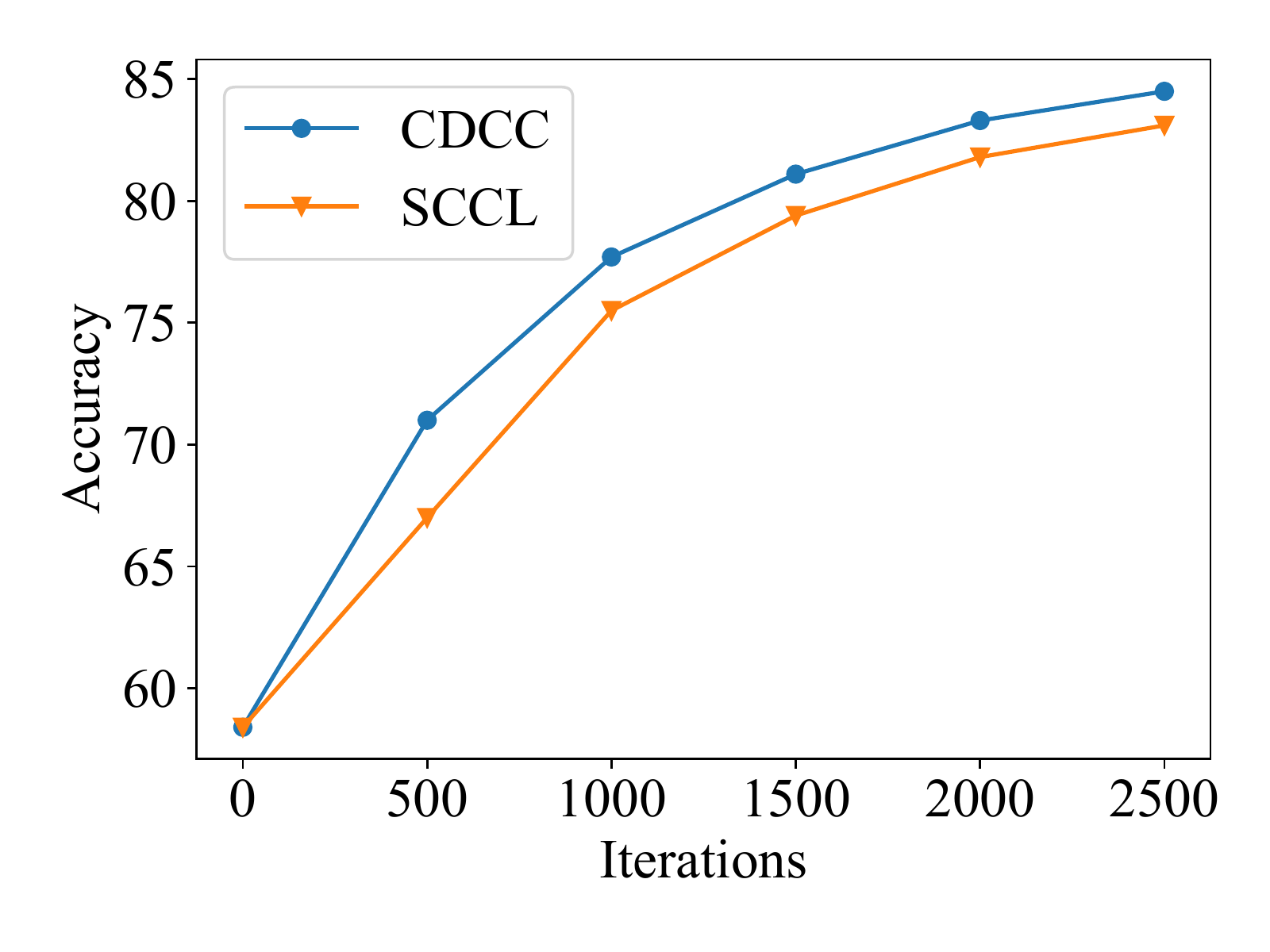}
  \caption{Top-1 Accuracy}
  \end{subfigure}

\caption{The GoogleNews-S results of CDCC and SCCL in terms of inter-cluster distances and top-1 accuracy during the learning of clustering.}
\label{fig:CategoryLoss}
\end{figure}

\begin{figure}[t]
\centering
\begin{subfigure}[b]{.8\columnwidth}
  \includegraphics[width=\columnwidth]{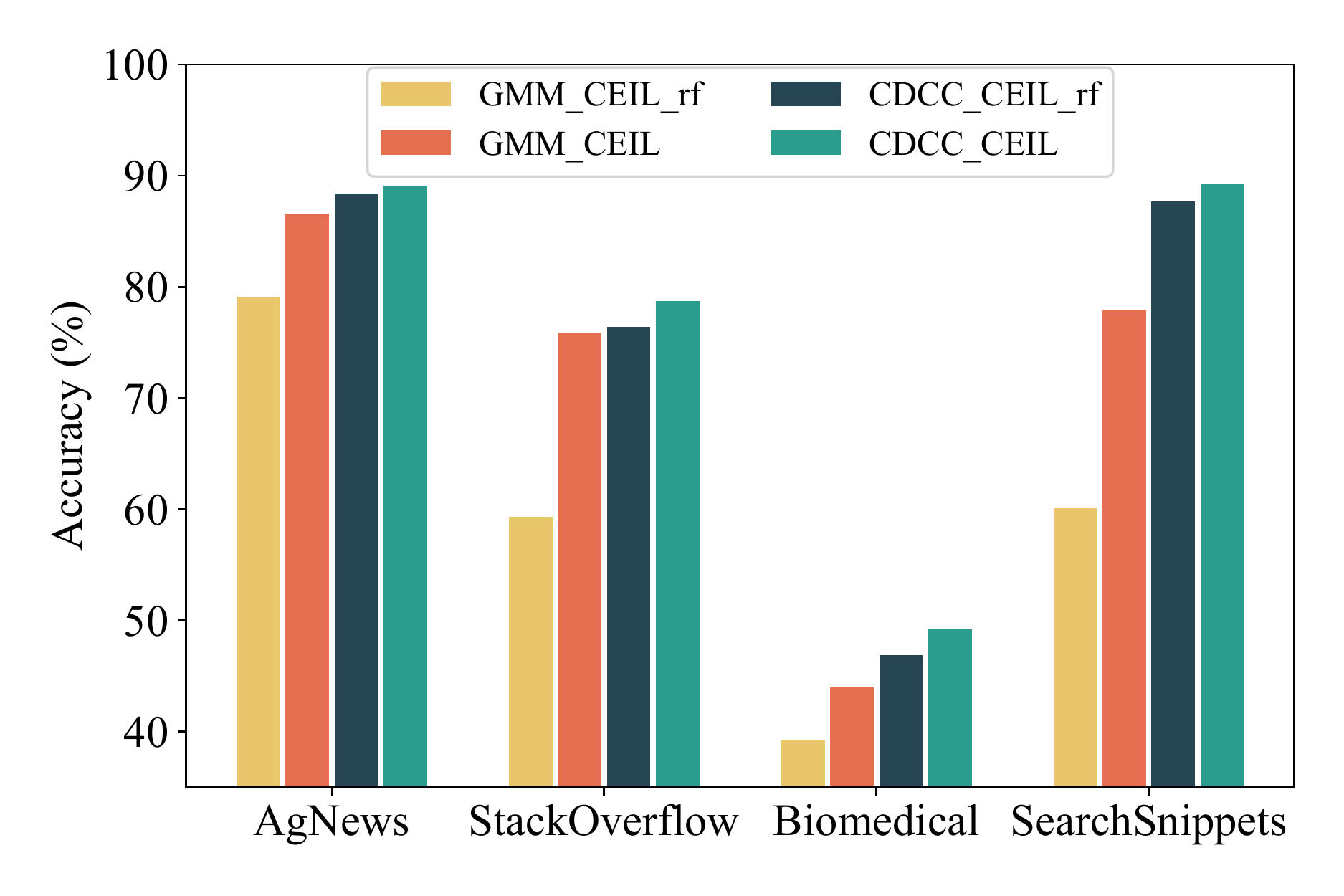}
  \caption{Results w/wo data filtering}
  \label{fig:filter}
\end{subfigure}
\begin{subfigure}[b]{\columnwidth}
  \includegraphics[width=\columnwidth]{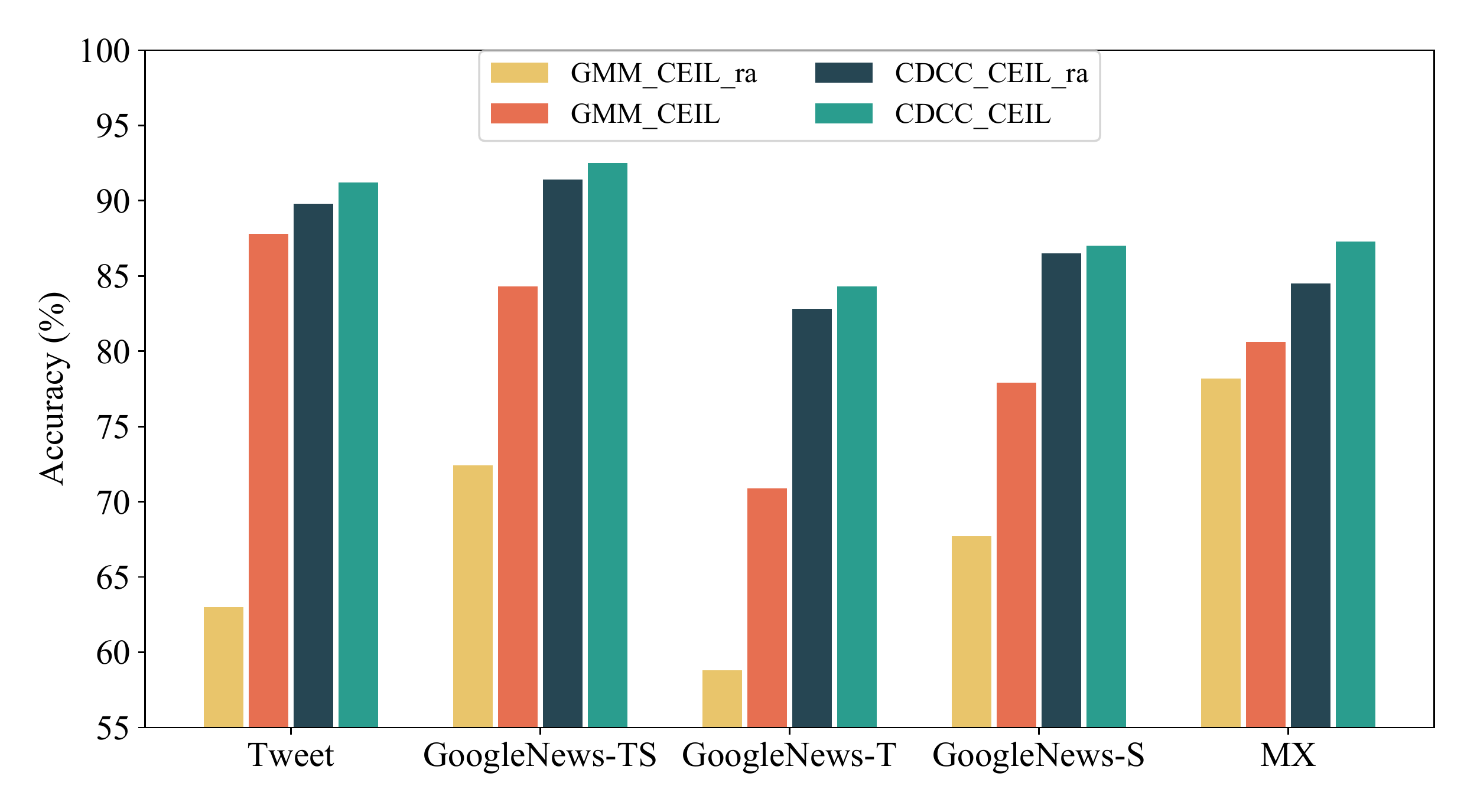}
  \caption{Results w/wo data aggregation}
  \label{fig:aggr}
\end{subfigure}
% \begin{subfigure}[b]{.32\columnwidth}
%   \includegraphics[width=\columnwidth]{figures/category_loss/(a)Googlenews-S_NMI.pdf}
%   \caption{ImageNet policy}
%   \label{fig:imgnet}
% \end{subfigure}

\caption{The ablative analysis of the data filtering and aggregation modules in \textit{CEIL} evaluated using GMM and CDCC algorithms. ``\_rf'' and ``\_ra'' denote the filtering and aggregation module is removed from the clustering framework respectively. }
\label{fig:FilterAggr}
\end{figure}

\begin{table*}[tb]
    \centering
    \caption{Comparison results of the prompt learning approach (``\_p'') and the fine-tuning approach (``\_f'') for feature extraction and classification learning.}
    \resizebox{\textwidth}{!}{
    \begin{tabular}{c|cc|cccccccccccccccccc}
        \toprule
        \multirow{2}{*}{Method} & \multicolumn{2}{c|}{Average} & \multicolumn{2}{c}{AgNews} & \multicolumn{2}{c}{StackOverflow} & \multicolumn{2}{c}{Biomedical} & \multicolumn{2}{c}{SearchSnippets} & \multicolumn{2}{c}{Tweet} & \multicolumn{2}{c}{GoogleNews-TS} & \multicolumn{2}{c}{GoogleNews-S} & \multicolumn{2}{c}{GoogleNews-T} & \multicolumn{2}{c}{MX} \\
         & ACC & NMI & ACC & NMI & ACC & NMI & ACC & NMI & ACC & NMI & ACC & NMI & ACC & NMI & ACC & NMI & ACC & NMI & ACC & NMI \\
        \midrule
        CDCC\_f & $78.8$ & $\mathbf{76.8}$ & $\mathbf{88.3}$ & $\mathbf{68.2}$ & $76.9$ & $\mathbf{75.3}$ & $42.8$ & $\mathbf{38.1}$ & $85.3$ & $71.2$ & $82.6$ & $90.2$ & $90.2$ & $94.9$ & $84.5$ & $90.6$ & $76.2$ & $\mathbf{88.5}$ & $82.8$ & $74.3$\\
        CDCC\_p & $\mathbf{79.9}$ & $76.7$ & $\mathbf{88.3}$ & $\mathbf{68.2}$ & $\mathbf{77.0}$ & $\mathbf{75.3}$ & $\mathbf{43.4}$ & $34.2$ & $\mathbf{88.0}$ & $\mathbf{72.3}$ & $\mathbf{86.8}$ & $\mathbf{91.9}$ & $\mathbf{90.8}$ & $\mathbf{95.0}$ & $\mathbf{84.8}$ & $\mathbf{90.8}$ & $\mathbf{77.1}$ & $87.7$ & $\mathbf{83.2}$ & $\mathbf{74.5}$ \\
        \midrule
        CDCC\_CEIL\_f & $81.2$ & $77.6$ & $88.5$ & $68.3$ & $77.6$ & $76.5$ & $47.5$ & $38.1$ & $88.2$ & $73.1$ & $88.2$ & $92.1$ & $91.2$ & $95.5$ & $86.8$ & $91.0$ & $78.5$ & $88.3$ & $84.1$ & $75.1$\\
        CDCC\_CEIL\_p & $\mathbf{83.2}$ & $\mathbf{79.2}$ & $\mathbf{89.1}$ & $\mathbf{69.7}$ & $\mathbf{78.8}$ & $\mathbf{77.3}$ & $\mathbf{49.2}$ & $\mathbf{41.0}$ & $\mathbf{89.3}$ & $\mathbf{75.2}$ & $\mathbf{91.2}$ & $\mathbf{94.1}$ & $\mathbf{92.5}$ & $\mathbf{95.8}$ & $\mathbf{87.2}$ & $\mathbf{92.4}$ & $\mathbf{84.3}$ & $\mathbf{91.3}$ & $\mathbf{87.3}$ & $\mathbf{76.1}$ \\
        \bottomrule
    \end{tabular}
    }
    \label{tab:Prompt}
\end{table*}

\subsection{Ablation Studies}
We have conducted comprehensive ablation analysis to evaluate the impact of different modules in our method, including the category loss in the proposed CDCC algorithm, the data filtering and aggregation module, the prompt learning approach for representation construction and classification learning, and the iterative updating of the \textit{CEIL} framework.

\subsubsection{Category Loss.}
While the main difference between CDCC and SCCL is the additional category loss incorporated in CDCC,
we have conducted experiments on GoogleNews-S to evaluate its effectiveness by comparing the inter-cluster distances and accuracy of CDCC and SCCL over iterations.

Figure \ref{fig:CategoryLoss} shows the comparison results, where the CDCC algorithm produces more separated clusters and constantly outperforms SCCL with higher accuracy by incorporating the proposed category loss.
This observation validates our hypothesis that the category loss achieves better separation between clusters, leading to more accurate clustering.

\subsubsection{Data Filtering and Aggregation.}
Experiments were conducted to assess the impact of the data filtering and aggregation module by separately removing them from the framework and evaluating the clustering performance.

In Figure \ref{fig:filter} and \ref{fig:aggr}, we demonstrate the comparison results of the complete \textit{CEIL} framework on GMM and CDCC with the frameworks where the data filtering or aggregation module is removed, denoted with the ``\_rf'' and ``\_ra'' suffixes respectively.
Results show that removing either the data filtering module or aggregation module will cause significant performance drop for GMM\_CEIL.
The reason is that in GMM\_CEIL, the representation learning is only performed at the update of the classification task.
Therefore, the quality of classification labels is critical for the clustering performance, which is greatly impacted by the data filtering and aggregation processes.
In CDCC\_CEIL, the performance decline of removing data filtering and aggregation modules is relatively small compared to GMM\_CEIL, as the CDCC algorithm itself learns good text representations.
% the data filtering and aggregation processes are critical for the quality of classification labels, hence having greatly impact on the final clustering performance.

\subsubsection{Prompt Learning.}
We adopt prompted input text segments for representation extraction, and prompt-tuning for the learning of the classifier.
To demonstrate its effectiveness, we compare it with the normal fine-tuning approach, which uses the original text segments as the input for representation extraction and classification, and adopts an MLP network as the classification head for fine-tuning.

Results in Table \ref{tab:Prompt} show that by using prompt learning, CDCC\_p and CDCC\_CEIL\_p both achieve generally better clustering results compared with their fine-tuning counterparts. 
% the prompt learning approach achieves generally better clustering results than the fine-tuning approach.
The success is contributed to the capability of prompt templates to better model the semantics when little supervision is provided.

\begin{figure}[t]
\centering
\begin{subfigure}[b]{.49\columnwidth}
\includegraphics[width=\columnwidth]{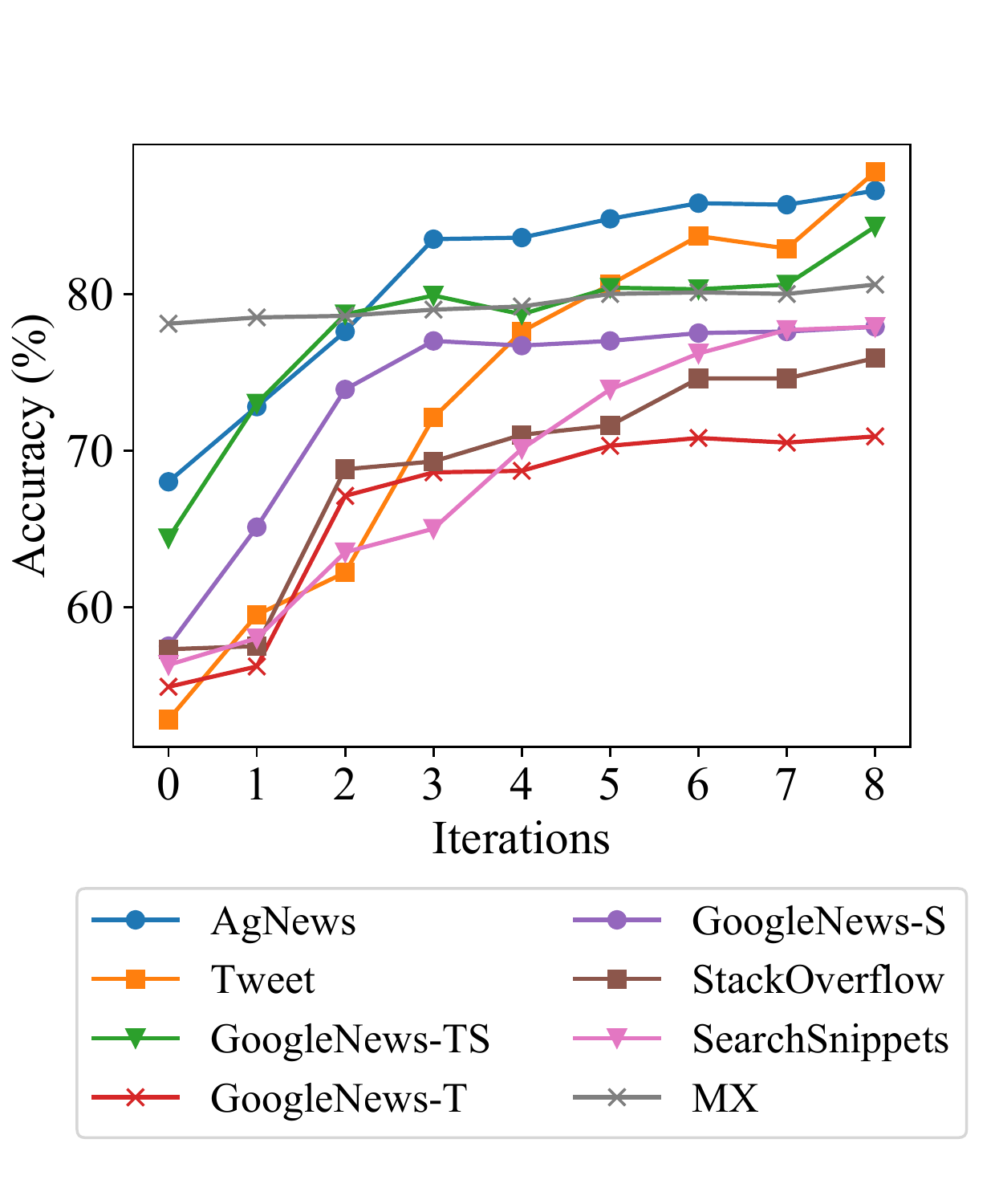}
  \caption{GMM\_CEIL results}
  \label{fig:iter_gmm}
\end{subfigure}
\begin{subfigure}[b]{.49\columnwidth}
  \includegraphics[width=\columnwidth]{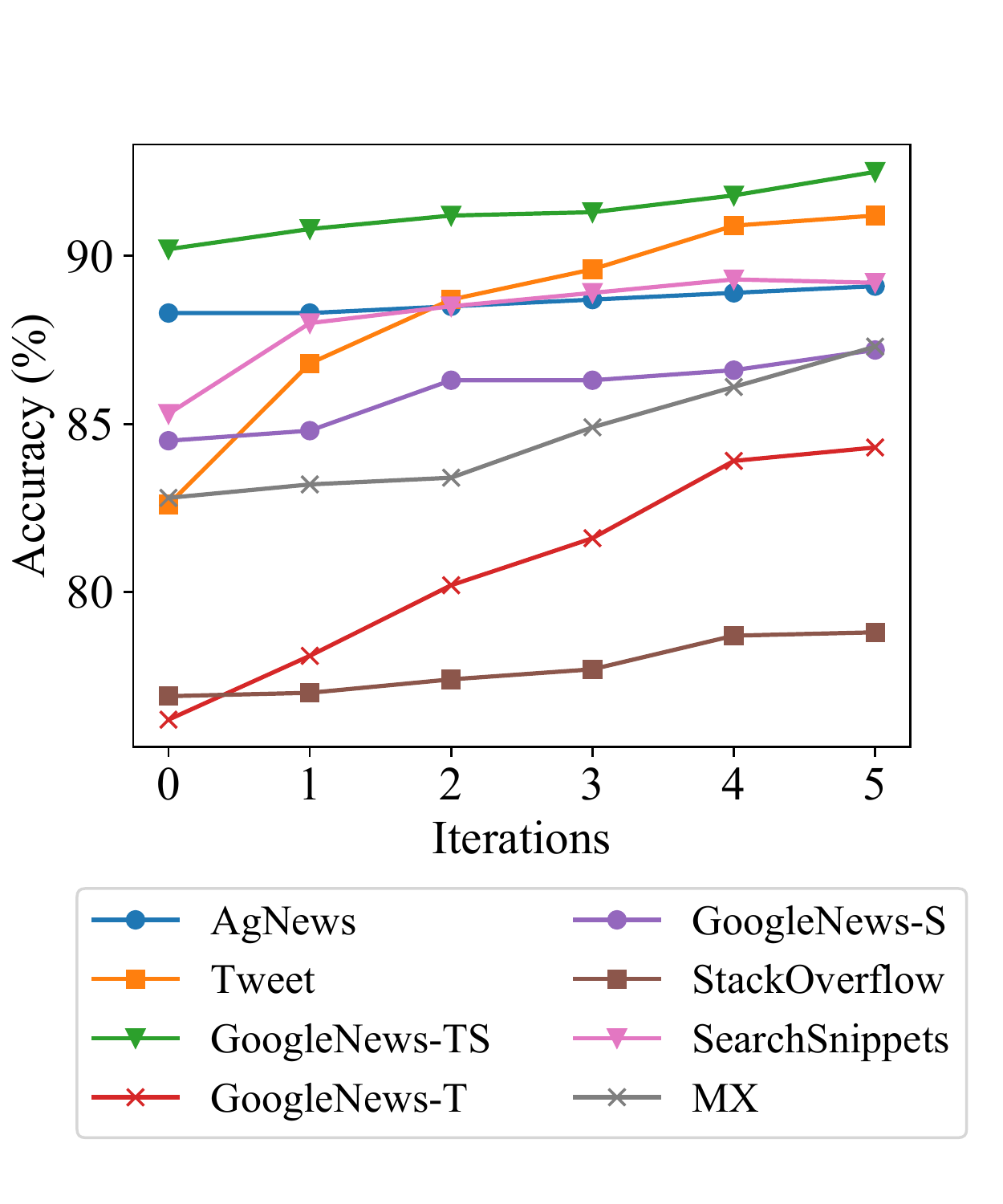}
  \caption{CDCC\_CEIL results}
  \label{fig:iter_cdcc}
\end{subfigure}

\caption{The clustering performance of the \textit{CEIL} framework with GMM and CDCC over different iterations on various short text clustering tasks.}
\label{fig:Stability}
\end{figure}

\subsubsection{Iterative Updating.}
In the \textit{CEIL} framework, the clustering task and the classification task are supposed to promote each other in each iteration, and gain better clustering results and text representations iteratively.
In order to verify this motivation, we conduct experiments on multiple short text clustering datasets and plot the performance of GMM\_CEIL and CDCC\_CEIL over iterations, shown in Figure \ref{fig:Stability}.

We can observe a rapid increase of clustering performance over iterations on GMM\_CEIL shown in Figure \ref{fig:iter_gmm}.
This is because our \textit{CEIL} framework can iteratively enhance the text representations based on the feedback of clustering results, leading to enormous improvements to traditional clustering algorithms like GMM which are highly dependent on the quality of data representations.
A clear trend of performance increase over iterations can also be observed in Figure \ref{fig:iter_cdcc}, when \textit{CEIL} is applied to powerful deep clustering algorithms like CDCC, which again illustrates the effectiveness and generality of our proposed \textit{CEIL} framework.

% there is a clear trend that the clustering performance increases with the number of iterations on all tasks.

\section{Conclusion}
In this paper, we propose a Classification-Enhanced Iterative Framework for short text clustering, where a classification objective is introduced in the clustering framework to iteratively enhance the text representations which further improves the clustering performance.
We also propose a Category Disentangled Contrastive Clustering algorithm which outperforms other clustering algorithms by achieving better separation among clusters.
Experimental results demonstrate that our \textit{CEIL} generally improves the performance of various clustering algorithms and achieves the state-of-the-art performance on a wide range of short text clustering tasks.

\bibliographystyle{ACM-Reference-Format}
\bibliography{main}

\end{document}